\documentclass[conference]{IEEEtran}
\IEEEoverridecommandlockouts

\makeatletter
\def\endthebibliography{%
	\def\@noitemerr{\@latex@warning{Empty `thebibliography' environment}}%
	\endlist
}
\makeatother
\usepackage{xcolor}
\usepackage{hyperref}
\hypersetup{draft}
\usepackage{verbatim}
\usepackage{amsmath}
\usepackage{textcomp}
\usepackage[pdftex]{graphicx}
\usepackage{multicol}
\usepackage{multirow}
\usepackage{caption}
\usepackage{subcaption}
\usepackage{array}
\usepackage{fixltx2e}
\usepackage{graphicx} 
\usepackage{tabularx} 
\usepackage{booktabs} 
\usepackage{longtable}
\usepackage{rotating}
\usepackage{adjustbox}
\begin{document}
\title{DemoBias: An Empirical Study to Trace Demographic Biases in Vision Foundation Models}

\author{Abu Sufian\\
CNR-ISASI\\ 73100 Lecce, Italy.\\
{\tt\small abusufian@cnr.it}
\and
Anirudha Ghosh\\
Visva-Bharati\\
Santiniketan, India\\
{\tt\small ghoshanirudha141@gmail.com}
\and
Debaditya Barman\\
Visva-Bharati\\
Santiniketan, India\\
{\tt\small debadityabarman@gmail.com}
\and
Marco Leo \\
CNR-ISASI\\ 73100 Lecce, Italy\\
{\tt\small marco.leo@cnr.it}
\and
Cosimo Distante\\
CNR-ISASI\\ 73100 Lecce, Italy\\
{\tt\small Cosimo.distante@cnr.it}
}
\maketitle
\begin{abstract}
Large Vision Language Models (LVLMs) have demonstrated remarkable capabilities across various downstream tasks, including biometric face recognition (FR) with description. However, demographic biases remain a critical concern in FR, as these foundation models often fail to perform equitably across diverse demographic groups, considering ethnicity/race, gender, and age. Therefore, through our work DemoBias, we conduct an empirical evaluation to investigate the extent of demographic biases in LVLMs for biometric FR with textual token generation tasks. We fine-tuned and evaluated three widely used pre-trained LVLMs: LLaVA, BLIP-2, and PaliGemma on our own generated demographic-balanced dataset. We utilize several evaluation metrics, like group-specific BERTScores and the Fairness Discrepancy Rate, to quantify and trace the performance disparities. The experimental results deliver compelling insights into the fairness and reliability of LVLMs across diverse demographic groups. Our empirical study uncovered demographic biases in LVLMs, with PaliGemma and LLaVA exhibiting higher disparities for Hispanic/Latino, Caucasian, and South Asian groups, whereas BLIP-2 demonstrated comparably consistent. Repository: https://github.com/Sufianlab/DemoBias.
\end{abstract}

\section{Introduction}
Modern Artificial Intelligence (AI) has transformed various sectors, driven by the advancements of Deep Learning (DL) \cite{lecun2015deep,prince2023understanding}, Transformer model \cite{vaswani2017attention,dosovitskiy2021an}, and large-scale datasets \cite{deng2009imagenet,parkhi2015deep}. This transformation reached its pinnacle with the advent of Large Language Models (LLMs) \cite{touvron2023llama,zhong2024evaluation,guo2025deepseek} and thereafter multimodal models like Large Vision Language Models (LVLMs) \cite{liu2024visual,li2023blip,beyer2024paligemma}, referred to as foundation models. These models have demonstrated huge success across a broad spectrum of downstream tasks, with biometric face recognition (FR) including textual description being one of the most prominent applications \cite{deandres2024good,zhang2024vision,awais2025foundation}. 

Despite the success of the foundation models, bias remains a significant concern in biometric FR. These models often exhibit disparities in performance across demographic groups, including ethnicity/race, gender, and age \cite{yucer2024racial,drozdowski2020demographic,navigli2023biases}. One of the key factors contributing to this bias is the demographic imbalance in training data, as global digital data often underrepresent certain groups. Since each foundation model may have been trained on a different subset of these datasets, inherent biases could have been introduced. Furthermore, differences in model architectures and training processes influence how features are captured and represented, potentially amplifying biases. As a result, pre-trained models are often not bias-free.  

Various efforts have been made for bias mitigation in biometric FR \cite{yucer2024racial,drozdowski2020demographic,terhorst2021comprehensive}. Kotwal and Marcel introduced DeFT, a transformer model focused on demographic fairness \cite{kotwal2022fairness}. Howard et al. proposed functional fairness criteria and GARBE to balance accuracy and fairness \cite{howard2022evaluating}. Challenges in achieving fairness, especially for intersectional demographic groups, were highlighted by Atzori et al. \cite{atzori2022explaining} and Kolberg et al. \cite{kolberg2024potential}. Neural architecture search methods \cite{dooley2024rethinking}  have shown promise in optimizing fairness and accuracy, while Fu et al. used statistical activation maps for bias explainability \cite{fu2022towards}. Metrics such as fairness discrepancy rate \cite{de2021fairness} and unsupervised fair score normalization \cite{terhorst2020post} improved bias evaluation by quantifying the biases. Data augmentation techniques were also enhanced demographic representation in some proposed works \cite{yucer2020exploring,yao2023study}. Despite these advancements, demographic biases in biometric FR remain a persistent challenge.

To address these challenges, we conducted an empirical study, \textbf{DemoBias}, to examine demographic biases in state-of-the-art (SOTA) pre-trained LVLMs for biometric face recognition. Specifically, we evaluated LLaVA \cite{liu2024visual}, BLIP-2 \cite{li2023blip}, and PaLI-Gemma \cite{beyer2024paligemma}.  
 As we not found any suitable datasets for this study, we developed a demographically balanced dataset for fine-tuning and testing. This given us a unique opportunity to examine how these pre-trained models behave on the new dataset. We quantified biases using customized group-specific BERTScores, along with the Fairness Discrepancy Rate (FDR). The results provided interesting insights into the biases present in the LVLMs. Our empirical study identified notable demographic biases in the LVLMs, with PaliGemma and LLaVA exhibited higher disparities in textual descriptions for Hispanic/Latino, South Asian, and Caucasian groups, while BLIP-2 demonstrated more consistency across demographic groups but was not entirely bias-free. 

The key contributions of this paper are as follows:
\begin{itemize}
    \item We conduct an empirical evaluation of three popular LVLMs: LLaVA, BLIP-2, and PaliGemma on the biometric FR task.
    \item We develop a demographically balanced dataset for fine-tuning and testing, addressing the lack of suitable datasets in the existing literature.
    \item We employ group-specific customized metrics, including BERT\_Precision, BERT\_Recall, BERT\_F1 Scores, and FDR, to trace and quantify the demographic biases.
    \item We present detailed experimental results, revealing compelling insights into the fairness and reliability of LVLMs.
    \end{itemize}



\section{Materials and Methodology}
\label{Method}

This empirical study investigates the performance of three popular LVLMs: LLaVA, BLIP-2, and PaliGemma on the biometric FR task. Fine-tuning and testing were conducted using our demographically balanced dataset, and the models were evaluated with tailored metrics. Below, we describe the dataset, models, and metrics used in this study.

\subsection{Dataset}
\label{Dataset}

There is a lack of a demographically balanced dataset with textual token embeddings suitable for this experiment in the literature. Therefore, we developed a new dataset designed to provide a comprehensive representation across critical demographic attributes, enabling robust fairness evaluations. 

\begin{figure*}[!ht]
  \centering
  \includegraphics[width=\linewidth, height=0.65\linewidth]{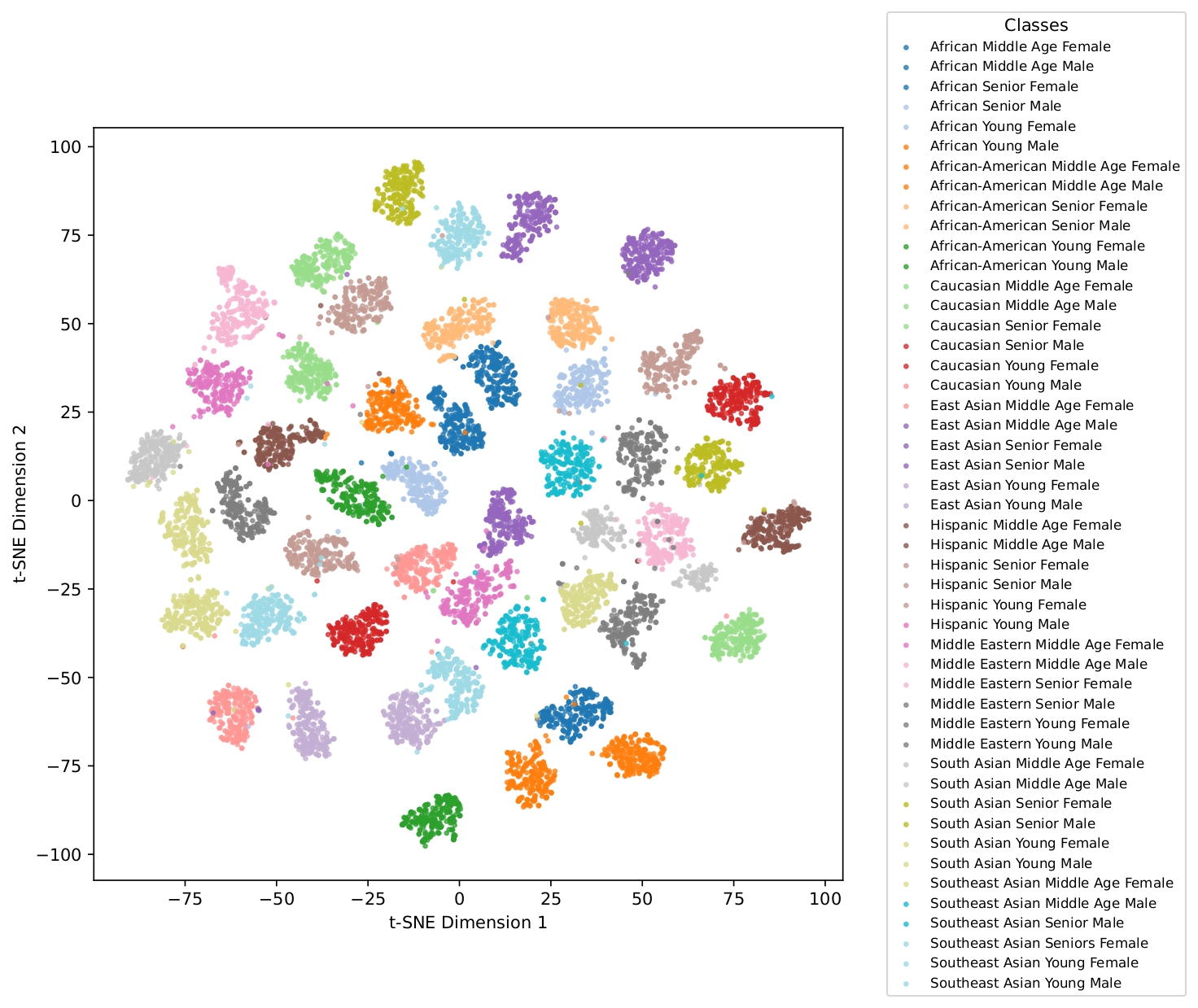}
  \caption{The t-SNE map visualizes the dataset's features after 100 epochs training using ResNet-18 \cite{he2016deep}. Each dot represent a face image, with different colors marking the 48 demographic groups. The formation of distinct clusters indicates distinguishing attributes for each group. Minor overlaps highlight that the dataset contains some demographically similar faces.}
  \label{fig:t-SNE}
\end{figure*}

The dataset consists of 12,000 face images of 240 public figures, categorized into 48 demographic groups based on ethnicity/race, gender, and age. The images were collected from multiple copyright-free sources and are intended solely for research on demographic bias detection and mitigation. Each image was resized to $224 \times 224$ pixels to balance computational cost and quality. The dataset was split in a 60:40 ratio for fine-tuning and testing, with detailed metadata in a JSON file capturing demographic attributes. This structure ensures the dataset is well-suited for fairness evaluation in biometric face recognition tasks. The dataset was organized into three primary demographic attributes: 
\begin{itemize}
    \item \textbf{Ethnicity/Race}: Eight distinct ethnic/racial groups, each comprising 30 individuals each of with 50 images, totaling 1,500 images per group. The groups are African, African-American, Caucasian, East Asian, Hispanic, Middle Eastern, South Asian, and Southeast Asian.  
    \item \textbf{Gender}: An equal representation of males and females within each ethnic/race group, with 120 individuals per gender, totaling 6,000 images per group.
    \item \textbf{Age Group}: We further divided the dataset into three age categories: Young (16–30 years), Middle Age (31–60 years), and Senior (60+ years). Each age group includes 80 individuals, totaling 4,000 images per group.
\end{itemize}

We have prioritized frontal-pose images during the image collection in order to minimize pose-related biases. Fig. \ref{fig:t-SNE} presents a t-SNE plot showing feature space extracted using ResNet-18 \cite{he2016deep} after 100 training epochs. Distinct clusters demonstrate the dataset's well structured representation with minimal bias.

\subsection{LLaVA}
\label{LLaVA}

LLaVA (Large Language and Vision Assistant) \cite{liu2024visual} integrates vision and language processing by combining a visual encoder with a LLM. It aligns image features with textual inputs, enabling multimodal reasoning and interaction. LLaVA's architecture typically includes a vision transformer for image encoding and a LLM (e.g., LLaMA \cite{touvron2023llama}) for textual token processing.
By mapping facial features to descriptive text, LLaVA enhances biometric face recognition tasks. Fine-tuning on datasets with image-text pairs allows LLaVA to perform context-aware recognition, handle unseen classes, and deliver adaptable performance in face biometric applications.

\subsection{BLIP-2}
\label{BLIP-2}

BLIP-2 is a LVLM  framework that leverages pre-trained image encoders and LLMs to reduce computational cost while achieving SOTA performance \cite{li2023blip}. It introduces a lightweight Querying Transformer (Q-Former) that bridges the modality gap between vision and language through a two-stage training approach. In the first stage, BLIP-2 performs vision-language representation learning by training the Q-Former to extract informative visual representations from a frozen image encoder. In the second stage, it enables vision-to-language generative learning by connecting the Q-Former to a frozen LLM, allowing it to generate textual outputs conditioned on visual inputs.
By utilizing frozen unimodal models, BLIP-2 reduces the number of trainable parameters. It achieves competitive results such as visual question answering, image captioning, and image-text retrieval, making it a valuable tool for various multimodal applications, including biometric FR.

\subsection{PaliGemma}
\label{PaliGemma}

PaliGemma (Pathways Language and Image model - Gemma) \cite{beyer2024paligemma} is a multimodal model combining a visual encoder and a LLM to align images and text in a unified representation space. This architecture excels in understanding and generating image-text pairs, making it well-suited for diverse computer vision tasks, including biometric face recognition.
Its multimodal learning approach maps facial features to descriptive text, enabling robust identity verification. Fine-tuned on image-text datasets, PaliGemma effectively adapts to domain-specific tasks and exhibits zero-shot recognition capabilities, making it a flexible and powerful tool for biometric applications.

\subsection{Metrics Used for Quantification of the Results}

We used following customized metrics to quantify and trace the demographic biases in the models.  

\subsubsection*{i) Group-Specific BERTScore}

Group-specific BERTScore trace the model's ability to capture semantic similarity between generated and embedded texts of each face image within each demographic group \( i \). For each group \( i \), we calculated BERT\_precision, BERT\_recall, and BERT\_F1 score based on text generation and text embeddings as in (\ref{eqn:BERT_Precision}), (\ref{eqn:BERT_Recall}), and (\ref{eqn:BERT_F1}).
\begin{align}
\label{eqn:BERT_Precision}
    \text{\footnotesize BERT\_Precision}_i & = \frac{1}{m_i} \sum_{j=1}^{m_i} \max_{k=1, \ldots, n_i} \text{sim}_{i}(c_j, r_k)\\
    \label{eqn:BERT_Recall}
    \text{\footnotesize BERT\_Recall}_i & =\frac{1}{n_i} \sum_{k=1}^{n_i} \max_{j=1, \ldots, m_i} \text{sim}_{i}(c_j, r_k)\\
    \label{eqn:BERT_F1}
    \text{\footnotesize BERT\_F1}_i & = 2 \times \frac{\text{\footnotesize BERT\_Precision}_i \times \text{\footnotesize BERT\_Recall}_i}{\text{\footnotesize BERT\_Precision}_i + \text{\footnotesize BERT\_Recall}_i}
\end{align}
Where \( \text{sim}_{i}(c_j, r_k) \) denotes the Cosine similarity between token embeddings of the \( j \)-th candidate token (i.e. $c_j$) and the \( k \)-th reference token (i.e. $r_k$) within group \( i \).
 \( m_i \) and \( n_i \) denote the number of tokens in the generation and embedded texts, respectively for group \( i \).

\subsubsection*{ii) Group-Specific Fairness Discrepancy Rate (FDR)}
We used group-specific FDR to measure variability in model performance by calculating the difference between the highest and lowest BERT\_F1 scores for text generation by LVLMs across demographic groups. The FDR was calculated as in (\ref{eqn:FDR}). This metric trace fairness by identifying performance variability across demographic groups. A lower FDR refers a fairer model behavior.
\begin{align}
\label{eqn:FDR}
    \text{\footnotesize FDR} &= \max(\text{\footnotesize BERT\_F1}_i) - \min(\text{\footnotesize BERT\_F1}_i)
\end{align}
Where \( \text{BERT\_F1}_i \) represents the BERT F1 score for each demographic group \( i \).


\begin{table*}[!ht]
\caption{Results derived from LLaVA \cite{liu2024visual}, BLIP-2 \cite{li2023blip}, and PaliGemma \cite{beyer2024paligemma} for textual description token generation, evaluated across 48 distinct demographic groups using BERTScores in biometric FR tasks. Each column represents an ethnicity/race, while gender and age groups are organized with sets of rows. }
    \centering
    \adjustbox{width=\textwidth}{
\begin{tabular}{lccccccccccccccccccccccccc}
\toprule
\large \textbf{\Large Metric} & 
\multicolumn{3}{c}{\textbf{\Large African}}& 
\multicolumn{3}{c}{\textbf{\Large African-American}} & 
\multicolumn{3}{c}{\textbf{\Large Caucasian}} & 
\multicolumn{3}{c}{\textbf{\Large Hispanic/Latino}} & 
\multicolumn{3}{c}{\textbf{\Large East Asian}} & 
\multicolumn{3}{c}{\textbf{\Large Middle Eastern}} &
\multicolumn{3}{c}{\textbf{\Large South Asian}} & 
\multicolumn{3}{c}{\textbf{\Large Southeast Asian}}  \\
\cmidrule(lr){2-4} \cmidrule(lr){5-7} \cmidrule(lr){8-10} \cmidrule(lr){11-13} \cmidrule(lr){14-16} \cmidrule(lr){17-19} \cmidrule(lr){20-22} \cmidrule(lr){23-25}
 & \textbf{\normalsize LLaVA} & \textbf{\normalsize BLIP-2} & \textbf{\normalsize PaliGemma} & \textbf{\normalsize LLaVA} & \textbf{\normalsize BLIP-2} & \textbf{\normalsize PaliGemma} & \textbf{\normalsize LLaVA} & \textbf{\normalsize BLIP-2} & \textbf{\normalsize PaliGemma} & \textbf{\normalsize LLaVA} & \textbf{\normalsize BLIP-2} & \textbf{\normalsize PaliGemma} & \textbf{\normalsize LLaVA} & \textbf{\normalsize BLIP-2} & \textbf{\normalsize PaliGemma} & \textbf{\normalsize LLaVA} & \textbf{\normalsize BLIP-2} & \textbf{\normalsize PaliGemma} & \textbf{\normalsize LLaVA} & \textbf{\normalsize BLIP-2} & \textbf{\normalsize PaliGemma} & \textbf{\normalsize LLaVA} & \textbf{\normalsize BLIP-2} & \textbf{\normalsize PaliGemma} \\
\midrule
\multicolumn{25}{c}{\textbf{\large Young Male}} \\
\midrule
\textbf{\large BERT\_Precision} & \large 0.781 & \large 0.828 & \large 0.759 & \large 0.837 & \large 0.820 & \large 0.892 & \large 0.497 & \large 0.574 & \large 0.500 & \large 0.533 & \large 0.652 & \large 0.661 & \large 0.704 & \large 0.753 & \large 0.730 & \large 0.560 & \large 0.629 & \large 0.598 & \large 0.759 & \large 0.752 & \large 0.546 & \large 0.691 & \large 0.676 & \large 0.603 \\ 
\textbf{\large BERT\_Recall} & \large 0.836 & \large 0.842 & \large 0.642 & \large 0.820 & \large 0.857 & \large 0.676 & \large 0.646 & \large 0.605 & \large 0.379 & \large 0.666 & \large 0.667 & \large 0.591 & \large 0.758 & \large 0.708 & \large 0.517 & \large 0.552 & \large 0.559 & \large 0.495 & \large 0.778 & \large 0.761 & \large 0.383 & \large 0.557 & \large 0.564 & \large 0.205 \\ 
\textbf{\large BERT\_F1 Score} & \large 0.808 & \large 0.835 & \large 0.701 & \large 0.828 & \large 0.838 & \large 0.783 & \large 0.570 & \large 0.589 & \large 0.439 & \large 0.599 & \large 0.659 & \large 0.626 & \large 0.731 & \large 0.730 & \large 0.622 & \large 0.557 & \large 0.594 & \large 0.547 & \large 0.768 & \large 0.756 & \large 0.464 & \large 0.623 & \large 0.615 & \large 0.396 \\ 
\midrule
\multicolumn{25}{c}{\textbf{\large Young Female}} \\
\midrule
\textbf{\large BERT\_Precision} & \large 0.785 & \large 0.835 & \large 0.761 & \large 0.798 & \large 0.817 & \large 0.883 & \large 0.537 & \large 0.594 & \large 0.527 & \large 0.561 & \large 0.683 & \large 0.764 & \large 0.681 & \large 0.748 & \large 0.647 & \large 0.586 & \large 0.619 & \large 0.633 & \large 0.762 & \large 0.763 & \large 0.630 & \large 0.708 & \large 0.687 & \large 0.713 \\ 
\textbf{\large BERT\_Recall} & \large 0.834 & \large 0.843 & \large 0.673 & \large 0.807 & \large 0.854 & \large 0.708 & \large 0.691 & \large 0.623 & \large 0.367 & \large 0.622 & \large 0.691 & \large 0.711 & \large 0.733 & \large 0.703 & \large 0.357 & \large 0.573 & \large 0.549 & \large 0.514 & \large 0.797 & \large 0.764 & \large 0.461 & \large 0.587 & \large 0.578 & \large 0.328 \\ 
\textbf{\large BERT\_F1 Score} & \large 0.810 & \large 0.839 & \large 0.718 & \large 0.803 & \large 0.835 & \large 0.795 & \large 0.613 & \large 0.608 & \large 0.443 & \large 0.592 & \large 0.687 & \large 0.737 & \large 0.707 & \large 0.725 & \large 0.496 & \large 0.580 & \large 0.583 & \large 0.573 & \large 0.779 & \large 0.764 & \large 0.545 & \large 0.647 & \large 0.629 & \large 0.513 \\ 
\midrule
\multicolumn{25}{c}{\textbf{\large Middle Age Male}} \\
\midrule
\textbf{\large BERT\_Precision} & \large 0.808 & \large 0.840 & \large 0.713 & \large 0.840 & \large 0.826 & \large 0.805 & \large 0.538 & \large 0.596 & \large 0.495 & \large 0.539 & \large 0.689 & \large 0.893 & \large 0.758 & \large 0.753 & \large 0.706 & \large 0.542 & \large 0.641 & \large 0.676 & \large 0.775 & \large 0.772 & \large 0.568 & \large 0.704 & \large 0.672 & \large 0.677 \\ 
\textbf{\large BERT\_Recall} & \large 0.812 & \large 0.848 & \large 0.565 & \large 0.789 & \large 0.864 & \large 0.601 & \large 0.659 & \large 0.628 & \large 0.394 & \large 0.597 & \large 0.709 & \large 0.779 & \large 0.758 & \large 0.718 & \large 0.488 & \large 0.540 & \large 0.570 & \large 0.514 & \large 0.809 & \large 0.774 & \large 0.398 & \large 0.545 & \large 0.524 & \large 0.236 \\ 
\textbf{\large BERT\_F1 Score} & \large 0.810 & \large 0.844 & \large 0.639 & \large 0.815 & \large 0.844 & \large 0.702 & \large 0.598 & \large 0.612 & \large 0.444 & \large 0.568 & \large 0.699 & \large 0.836 & \large 0.758 & \large 0.735 & \large 0.595 & \large 0.541 & \large 0.605 & \large 0.594 & \large 0.792 & \large 0.773 & \large 0.483 & \large 0.623 & \large 0.594 & \large 0.448 \\ 
\midrule
\multicolumn{25}{c}{\textbf{\large Middle Age Female}} \\
\midrule
\textbf{\large BERT\_Precision} & \large 0.817 & \large 0.847 & \large 0.753 & \large 0.780 & \large 0.821 & \large 0.816 & \large 0.584 & \large 0.615 & \large 0.593 & \large 0.459 & \large 0.671 & \large 0.783 & \large 0.774 & \large 0.756 & \large 0.780 & \large 0.580 & \large 0.648 & \large 0.735 & \large 0.794 & \large 0.784 & \large 0.565 & \large 0.701 & \large 0.673 & \large 0.719 \\ 
\textbf{\large BERT\_Recall} & \large 0.817 & \large 0.849 & \large 0.639 & \large 0.756 & \large 0.857 & \large 0.619 & \large 0.680 & \large 0.648 & \large 0.395 & \large 0.443 & \large 0.689 & \large 0.632 & \large 0.779 & \large 0.729 & \large 0.563 & \large 0.584 & \large 0.580 & \large 0.555 & \large 0.822 & \large 0.790 & \large 0.367 & \large 0.547 & \large 0.537 & \large 0.328 \\ 
\textbf{\large BERT\_F1 Score} & \large 0.817 & \large 0.848 & \large 0.696 & \large 0.768 & \large 0.838 & \large 0.716 & \large 0.632 & \large 0.631 & \large 0.491 & \large 0.452 & \large 0.680 & \large 0.706 & \large 0.776 & \large 0.742 & \large 0.669 & \large 0.582 & \large 0.613 & \large 0.644 & \large 0.808 & \large 0.787 & \large 0.465 & \large 0.623 & \large 0.601 & \large 0.516 \\ 
\midrule
\multicolumn{25}{c}{\textbf{\large Senior Male}} \\
\midrule
\textbf{\large BERT\_Precision} & \large 0.815 & \large 0.836 & \large 0.601 & \large 0.806 & \large 0.832 & \large 0.851 & \large 0.574 & \large 0.583 & \large 0.402 & \large 0.584 & \large 0.682 & \large 0.804 & \large 0.739 & \large 0.750 & \large 0.572 & \large 0.535 & \large 0.599 & \large 0.585 & \large 0.583 & \large 0.776 & \large 0.402 & \large 0.656 & \large 0.681 & \large 0.514 \\ 
\textbf{\large BERT\_Recall} & \large 0.864 & \large 0.843 & \large 0.482 & \large 0.797 & \large 0.862 & \large 0.644 & \large 0.690 & \large 0.609 & \large 0.177 & \large 0.657 & \large 0.693 & \large 0.673 & \large 0.791 & \large 0.760 & \large 0.326 & \large 0.577 & \large 0.555 & \large 0.457 & \large 0.704 & \large 0.779 & \large 0.276 & \large 0.510 & \large 0.537 & \large 0.146 \\ 
\textbf{\large BERT\_F1 Score} & \large 0.839 & \large 0.839 & \large 0.541 & \large 0.801 & \large 0.847 & \large 0.746 & \large 0.632 & \large 0.596 & \large 0.283 & \large 0.620 & \large 0.687 & \large 0.737 & \large 0.765 & \large 0.755 & \large 0.445 & \large 0.556 & \large 0.577 & \large 0.521 & \large 0.643 & \large 0.777 & \large 0.338 & \large 0.582 & \large 0.601 & \large 0.323 \\ 
\midrule
\multicolumn{25}{c}{\textbf{\large Senior Female}} \\
\midrule
\textbf{\large BERT\_Precision} & \large 0.799 & \large 0.831 & \large 0.651 & \large 0.835 & \large 0.828 & \large 0.659 & \large 0.668 & \large 0.609 & \large 0.411 & \large 0.573 & \large 0.681 & \large 0.671 & \large 0.734 & \large 0.747 & \large 0.608 & \large 0.596 & \large 0.635 & \large 0.679 & \large 0.757 & \large 0.763 & \large 0.373 & \large 0.649 & \large 0.677 & \large 0.513 \\ 
\textbf{\large BERT\_Recall} & \large 0.845 & \large 0.839 & \large 0.545 & \large 0.820 & \large 0.859 & \large 0.526 & \large 0.769 & \large 0.643 & \large 0.158 & \large 0.608 & \large 0.689 & \large 0.450 & \large 0.767 & \large 0.761 & \large 0.347 & \large 0.608 & \large 0.558 & \large 0.464 & \large 0.809 & \large 0.781 & \large 0.181 & \large 0.497 & \large 0.520 & \large 0.164 \\ 
\textbf{\large BERT\_F1 Score} & \large 0.822 & \large 0.835 & \large 0.598 & \large 0.828 & \large 0.843 & \large 0.593 & \large 0.718 & \large 0.626 & \large 0.279 & \large 0.590 & \large 0.685 & \large 0.558 & \large 0.750 & \large 0.754 & \large 0.474 & \large 0.602 & \large 0.596 & \large 0.569 & \large 0.783 & \large 0.772 & \large 0.274 & \large 0.572 & \large 0.592 & \large 0.333 \\ 

\bottomrule
\end{tabular}}
\label{tab:Metrics}
\end{table*}

\section{Experiments and Results}  
\label{Result}
\subsection{System Configuration and Experiment Setup}
We conducted experiments on a multi-GPU workstation equipped with three 24 GB NVIDIA TITAN RTX GPUs with 62 GB of VRAM. For efficient training, we used DeepSpeed’s ZeRO-3 framework with LoRA for memory management and parameter updates, using FP16 precision for fine-tuning the LLaVA model.
Additionally, we used LoRA to update only specific parts (query projection (q\_proj) and key projection (k\_proj) layers) of the PaliGemma and BLIP-2  models. These models were quantized to 8-bit to optimize memory use and fine-tuned using the Adam optimizer with a Cosine Learning Rate (LR) schedule applied over iterations. Minimal fine-tuning was performed with a limited number of epochs.

For the dataset, approximately 60\% of each class sample was used for fine-tuning, while the remaining samples were reserved for testing. During training, both questions and predictions were performed in half-precision (float16) to accelerate the testing process.

The dataset for fine-tuning LVLMs was preprocessed as follows: images were converted to JPEG format and assigned unique IDs based on ethnicity, gender, age, and individual identity. They were paired with structured textual token detailing demographic attributes (e.g., ``An African Person, African Female, African Middle-Aged Female, Person AMAF\_1'') to create image-text embeddings. The split training and testing datasets were saved in JSON format, with each JSON file structured to use a single, consistent prompt (``Identify the person and demographics for this image.'') for all images.

All three modes were fine-tuned then models' output were evaluated using BERTScores. E.g., with prediction: ``A Caucasian Person, Caucasian Female, Caucasian Senior Female, Person CFXXF\_1" and the ground truth: ``A Caucasian Person, Caucasian Female, Caucasian Senior Female, Person CFXXF\_10," here demographic prediction was okay but missed to identity the person. The BERTScore measures alignment, indicating how closely the model’s output matches expected demographic attributes. This approach provides a quantitative assessment of semantic consistency between predicted and embedded tokens.

\subsection{Results}
 
This section presents the experimental results. Table \ref{tab:Metrics} summarizes the evaluation performance of LLaVA, BLIP-2, and PaliGemma, demonstrating these model's effectiveness in identifying faces and generating descriptive tokens after fine-tuning. These results provide insights into the performance of SOTA LVLMs across different demographic groups in the context of biometric FR. While these findings may not be sufficient for drawing any generalizable conclusions, they underscore the presence of demographic biases in pre-trained LVLMs, as discussed in the following subsections.

\subsubsection{\textbf{Ethnicity/Race-Level Results}}
In Table \ref{tab:Metrics}, we present the results obtained from the three LVLM models: LLaVA, BLIP-2, and PaliGemma. For each ethnicity/race, organized by columns, we report BERT\_precision, BERT\_recall, and BERT\_F1 scores capturing the semantic similarity between the generated tokens and embedded text descriptions by the LVLM models.

\begin{figure*}[!ht]
\centering
\includegraphics[width=\linewidth, height=0.27\linewidth]{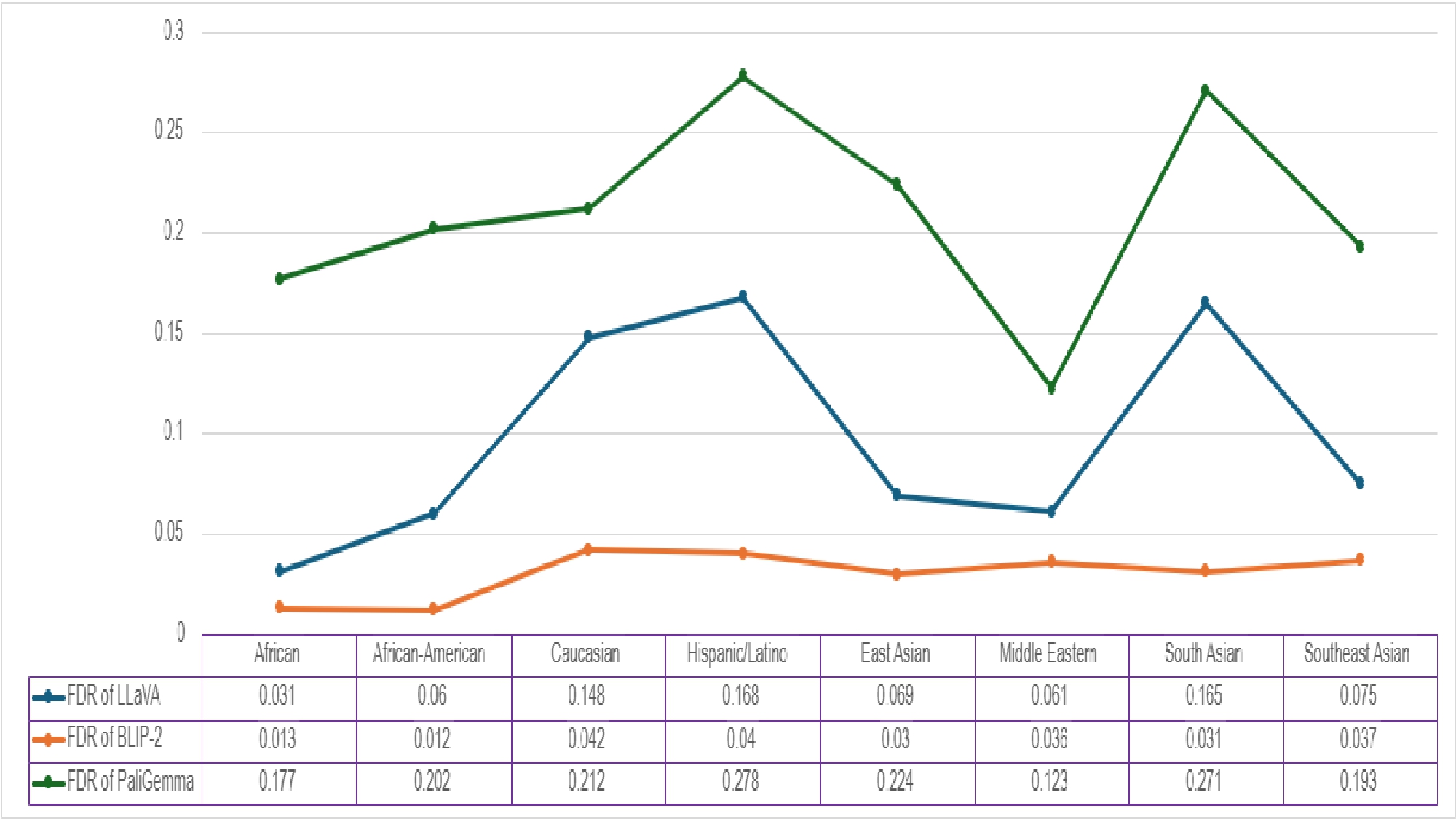}
\caption{This graph shows the Fairness of FDR of BERT\_F1 scores across eight ethnicities/races in all three models. A higher FDR value denotes a greater discrepancy, whereas the deviation of lines indicates that LVLM models did not exhibit demographic consistency in our empirical studies on our dataset.}
\label{fig:FDR1}
\end{figure*}

Fig. \ref{fig:FDR1} presents the FDR values returned by the LLaVA, BLIP-2, and PaliGemma models across various demographic groups. The FDR values have been determined using (\ref{eqn:FDR}). Our analysis reveals that the PaliGemma and LLaVA models introduce a higher demographic bias in biometric FR with textual descriptions, particularly for Hispanic/Latino (0.278 for PaliGemma, 0.168 for LLaVA), Caucasian (0.212 for PaliGemma, 0.148 for LLaVA), and South Asian (0.271 for PaliGemma, 0.165 for LLaVA) groups, indicating significant performance discrepancies across demographic groups. The BLIP-2 model exhibited lower demographic bias, with lower FDR values across most of the demographic groups. However, bias remains persistent for specific groups, including Caucasian, Hispanic, Southeast Asian, and Middle Eastern demographic groups. 

\subsubsection{\textbf{Age Group-Level Results}}
We evaluated performance of the three above-mentioned LVLM models across different age groups within each ethnicity/race. The results are presented in Table \ref{tab:Metrics}. By analyzing the results, we observe subtle age-related biases within certain ethnic/racial groups, posing challenges for the LVLMs. 

\begin{figure}[!ht]
\centering
\includegraphics[width=\linewidth]{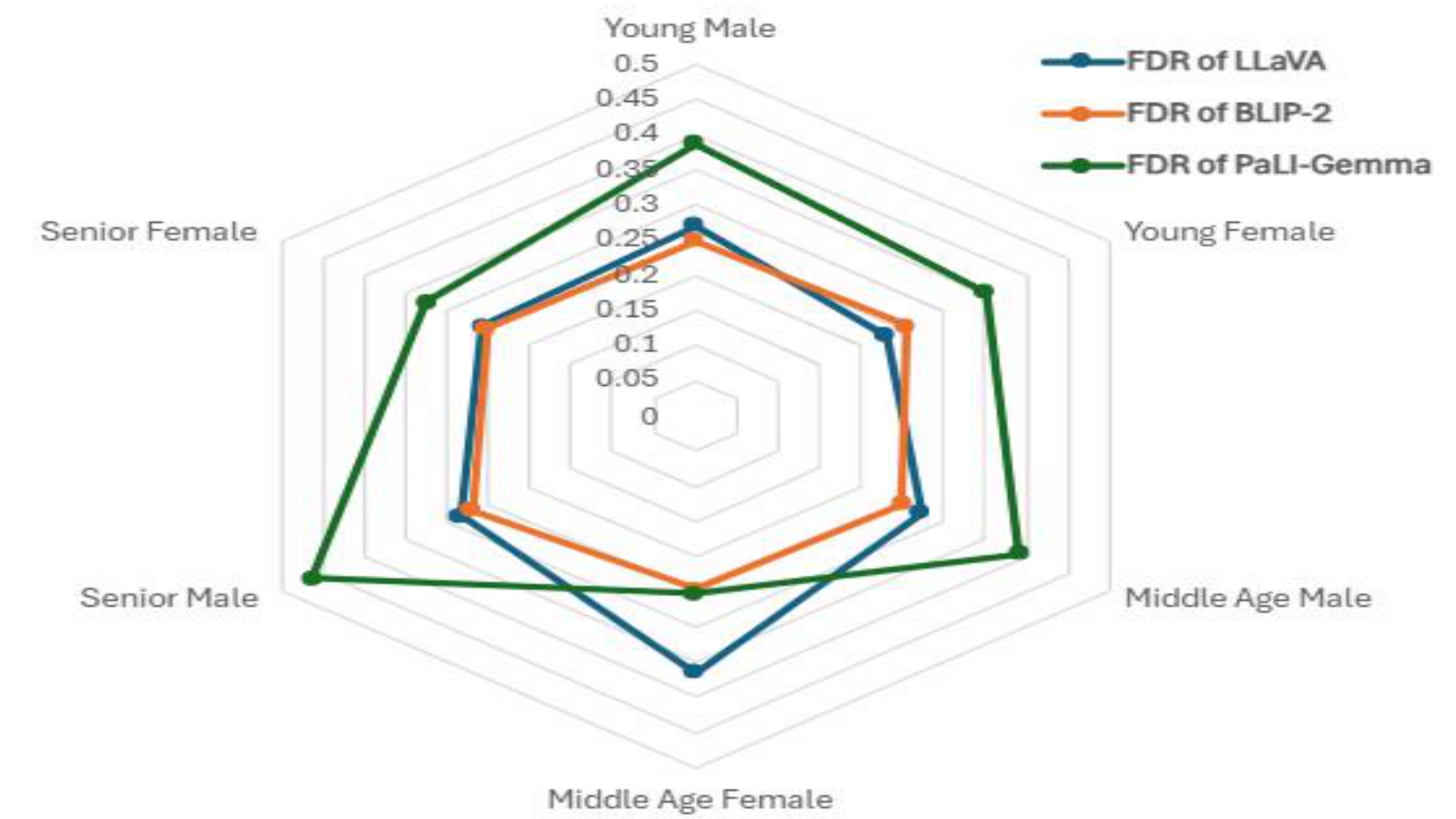}
\caption{This radar graph visualizes the FDR of BERT\_F1 scores across different age group images for all three models. Among them, BLIP-2 demonstrates the highest consistency.}
\label{fig:FDR2}
\end{figure}

The graph in Fig. \ref{fig:FDR2} illustrates the FDR results by different age groups. The FDR values indicate performance variability among these age groups, with Young Males (0.271 for LLaVA, 0.249 for BLIP-2, 0.387 for PaliGemma) and Middle-Aged Males (0.274 for LLaVA, 0.250 for BLIP-2, 0.392 for PaliGemma) exhibiting higher discrepancies. Middle-Aged Females (0.365 for LLaVA, 0.247 for BLIP-2, 0.252 for PaliGemma) and Senior Males (0.283 for LLaVA, 0.270 for BLIP-2, 0.463 for PaliGemma) also exhibited notable bias. These findings suggest that while BLIP-2 demonstrated greater consistency, PaliGemma exhibited higher bias across multiple age groups, and LLaVA exhibited moderate variability.

\subsubsection{\textbf{Gender-Level Results}}
We evaluated performance of the LVLM models from the perspective of gender within and across ethnicity/race groups. The results presented in Table \ref{tab:Metrics}, are organized by demographic group, with different set of rows representing distinct categories.

\begin{figure}[!ht]
\centering
\includegraphics[width=\linewidth, height=.43\linewidth]{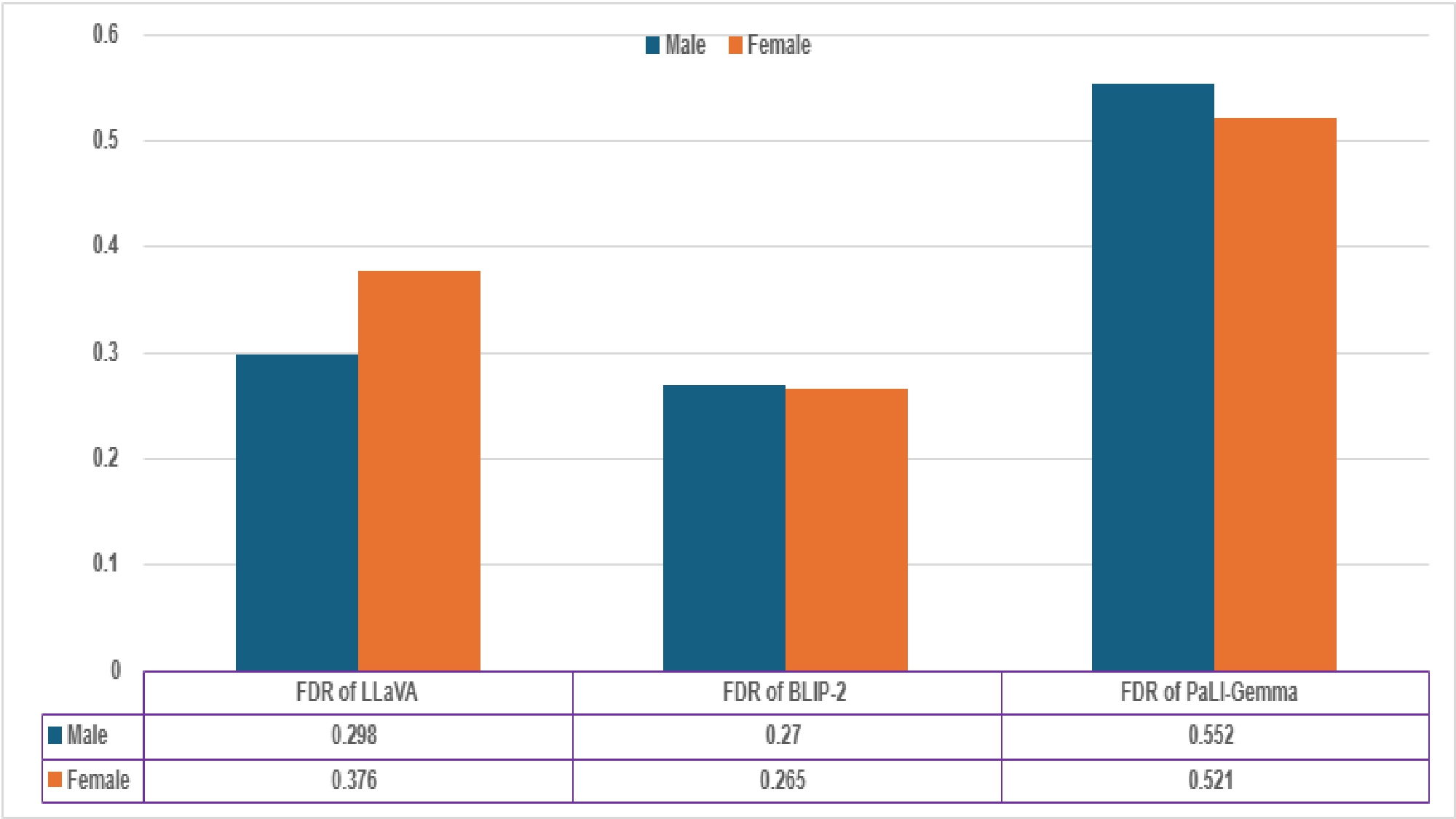}
\caption{This bar graph shows the fairness of FDR of BERT\_F1 scores of gender division in all three models. The varying heights of the bars indicate that the LVLM models did not exhibit gender-wise consistency in our empirical study.}
\label{fig:FDR3}
\end{figure}

Our analysis reveals gender biases within certain ethnic groups, particularly among closely related ones (e.g., Southeast Asian and East Asian). Fig. \ref{fig:FDR3} presents gender-wise FDR values, revealing that females (0.376 for LLaVA, 0.265 for BLIP-2, 0.521 for PaliGemma) exhibited little more discrepancies compared to males (0.298 for LLaVA, 0.270 for BLIP-2, 0.552 for PaliGemma). These results indicate that PaliGemma exhibited the highest gender bias, whereas BLIP-2 showed greater consistency across gender groups.

\section{Discussion and Future Scopes}\label{Discussion}

Our experimental results provide foundational insights into demographic fairness in LVLMs, highlighting areas where models like LLaVA, BLIP-2, and PaliGemma exhibited both strengths and biases. While BLIP-2 and LLaVA generally exhibited consistent performance across demographic groups, notable biases persisted, particularly among African and Hispanic/Latino groups. Key findings include: 
\\
\textbf{LLaVA} showed that it retained biases in Hispanic/Latino, South Asian, and Caucasian ethnicity categories. Differentiation across age groups were also noted along with gender-related biases, with higher variability among males.  \\
\textbf{BLIP-2} demonstrated better consistency, possibly due to its few-shot inference capability, but faced challenges with closely related ethnic/racial groups, such as Caucasian and Hispanic/Latino. Additionaly, it exhibited minor age-related variability, with consistency slightly decreasing from younger to senior groups.\\
\textbf{PaliGemma} demonstrated differentiation across demographic groups, retaining high biases in the Caucasian and Hispanic/Latino ethnic/racial groups but exhibiting less gender-related bias compared to LLaVA. Although it maintained semantic similarity with ground truth tokens, it often lacked exact word or phrase matches.

These findings establish a foundation for future research aimed at improving fairness in biometric FR. By highlighting demographic biases, this study underscores the importance of fairness-focused foundation model training. While our findings may not be sufficient to draw generalized conclusions, they clearly indicate that creating demographically balanced datasets, even for fine-tuning, can help to mitigate biases in LVLMs. Further empirical studies, refinement of fairness metrics, and development of domain adaptation techniques are crucial for enhancing generalization across diverse demographic groups.

\section{Conclusion}
\label{conclusion}

This study conducts an empirical analysis of demographic biases in LVLMs, specifically examining three state-of-the-art models: LLaVA, BLIP-2, and PaliGemma. By evaluating authentication and textual token generation for biometric face recognition, we identified persistent biases in pre-trained LVLM models, particularly across ethnic/racial and age groups. Our findings highlight the critical need for improved generalization and fairness in LVLMs to address demographic disparities in face biometric tasks.

Our work underscores the importance of developing equitable and fair AI systems that perform consistently across diverse demographic groups. By addressing these biases, we contribute to fostering more inclusive and fair AI, paving the way for responsible deployment in real-world scenarios.

\section*{Acknowledgment}
This research was partially supported by the project Future Artificial Intelligence Research FAIR CUP B53C220036 30006 grant number PE0000013. \\ \\
The authors thank Mr. Arturo Argentieri from CNR-ISASI, Italy, for his technical contributions to the multi-GPU computing facilities.

\bibliographystyle{IEEEtran}

\bibliography{OurBib}

\end{document}